\documentclass[11pt]{article}

\usepackage[margin=1in]{geometry}
\usepackage{times}
\usepackage{microtype}
\usepackage{float}
\usepackage{graphicx}
\usepackage{booktabs}
\usepackage{subcaption}
\usepackage{placeins}
\usepackage{amsmath,amssymb,amsthm}
\usepackage{natbib}
\usepackage{hyperref}
\usepackage{xcolor}

\usepackage{amsmath,amsfonts,bm}

\def\eqref#1{equation~\ref{#1}}

\def\1{\bm{1}}

\def\vb{{\bm{b}}}

\def\vu{{\bm{u}}}

\def\mW{{\bm{W}}}

\def\mPhi{{\bm{\Phi}}}

\DeclareMathAlphabet{\mathsfit}{\encodingdefault}{\sfdefault}{m}{sl}
\SetMathAlphabet{\mathsfit}{bold}{\encodingdefault}{\sfdefault}{bx}{n}

\newcommand{\softplus}{\zeta}

\newcommand{\LamICPath}[1]{figures/#1}

\newcommand{\LamICPlot}[3]{%
  \includegraphics[width=#3]{\LamICPath{#1}/#2.png}%
}

\title{Gradient Scaling Effects in Adaptive Spectral PINNs for Stiff Nonlinear ODEs}

\author{
Isabela M. Yepes, \ Pavlos Protopapas \\
Harvard University \\
\texttt{isabela\_yepes@g.harvard.edu, pprotopapas@g.harvard.edu}
}

\date{March 5, 2026}

\begin{document}

\maketitle

\begin{abstract}
Physics-Informed Neural Networks (PINNs) often struggle to train reliably on
stiff and oscillatory dynamical systems due to poor optimization conditioning.
While prior work has emphasized representational remedies such as spectral
parameterizations, the optimization implications of initial-condition (IC)
embeddings in adaptive spectral PINNs have not been well characterized.
In this work, we show that the choice of IC gating function induces explicit
time-dependent gradient scaling, which interacts with spectral representations
during training. Using a nonlinear stiff spring--pendulum ODE as a controlled
benchmark, we compare exponential and linear IC gates in combination with fixed
and adaptive Fourier spectral trunks. We observe stiffness-dependent changes in 
relative dominance for adaptive PINNs: at moderate stiffness ($k=20$), exponential gating often 
yields lower error but exhibits heterogeneous behavior across random seeds, 
whereas at higher stiffness ($k=60$), linear gating becomes preferable, with 
additional reversals observed at larger $k$. These trends hold for both relative 
$L^2$ error and maximum pointwise error and are confirmed by paired Wilcoxon 
signed-rank tests with Holm correction. Overall, our results demonstrate that IC 
embeddings are not a neutral design choice in PINNs: the induced gradient scaling 
materially shapes optimization conditioning in stiff regimes, with distinct sensitivity 
patterns in baseline and adaptive spectral models.
\end{abstract}

\section{Introduction}

Physics-Informed Neural Networks (PINNs) provide a flexible, mesh-free framework
for solving differential equations by embedding known physics into the training
objective. Despite their conceptual appeal, PINNs often struggle in stiff and
highly oscillatory regimes, where optimization becomes unstable or converges to
inaccurate solutions even when the governing equations are enforced exactly.

A central contributor to this behavior is spectral bias: standard neural
networks tend to learn low-frequency components of a target function much more
rapidly than high-frequency components. As stiffness increases and multiple time
scales emerge, the directions in function space corresponding to fast
oscillations become increasingly poorly conditioned under gradient-based
optimization. This phenomenon has been analyzed through optimization and kernel
perspectives, including Neural Tangent Kernel (NTK) analyses showing severe
eigenvalue imbalance and effective rank collapse in physics-informed settings
\citep{jacot2018ntk,wang2021whyfail}.

Representation-level remedies have therefore received significant attention.
Fourier feature networks reshape the effective kernel spectrum seen by gradient
descent and accelerate the learning of oscillatory components
\citep{wang2020fourier}, while more recent Separated-Variable Spectral Neural
Networks (SV-SNN) introduce explicit spectral structure to improve conditioning
and prevent rank collapse in high-frequency PDEs \citep{xiong2025svsnn}. These
approaches demonstrate that representation choice plays a critical role in
stabilizing optimization in stiff regimes.

In this work, we adopt a complementary optimization-driven perspective and
identify an additional, often underemphasized design choice that strongly
influences training dynamics: the parameterization used to enforce initial
conditions. The use of output transformations to enforce initial or boundary
conditions dates back to \citet{lagaris1998artificial} and is widely used in PINNs
(e.g., \cite{babni2025error}). While typically viewed as representationally
neutral, we make explicit that IC gating induces time-dependent Jacobian scaling.
Under linearization, gradient descent dynamics are governed by Jacobian inner
products (NTK), and the IC gate rescales these Jacobians
across the physical ODE trajectory $t$, thereby modifying optimization conditioning 
without changing the underlying function class defined by the IC constraint.

Using a nonlinear stiff spring--pendulum ODE as a controlled benchmark, we
systematically compare exponential and linear IC gates in combination with fixed
and adaptive Fourier spectral trunks. We observe stiffness-dependent changes in 
relative dominance: exponential gating performs better at moderate stiffness, 
whereas linear gating becomes preferable at higher stiffness levels, with 
additional reversals observed at larger $k$. Our study is intentionally diagnostic: 
by isolating IC gating under fixed optimization settings, we expose how this 
architectural choice reshapes conditioning in stiff regimes.

\section{Methods}

We study the planar spring--pendulum system in polar coordinates
$\vu(t)=[r(t),\theta(t)]^\top$, governed by the second-order ODE
\begin{equation}
\begin{alignedat}{2}
\ddot r \; &=\; r \dot{\theta}^2
- \frac{k}{m}(r - L_0)
+ g \cos\theta
- \frac{c_r}{m}\dot r,
\qquad
&
\ddot\theta \; &=\;
- \frac{2 \dot r \dot\theta}{r}
- \frac{g}{r}\sin\theta
- c_\theta \dot\theta .
\end{alignedat}
\end{equation}
The spring constant $k$ controls the stiffness of the system. 
All experiments are conducted over a fixed time horizon $T=10$, with
physical parameters $g=9.81$, $m=1.0$, $L_0=1.0$, $c_r=c_\theta=0$,
and $r_{\min}=10^{-4}$. 
High-accuracy reference solutions are generated using
\texttt{solve\_ivp} with the \texttt{DOP853} integrator and strict
tolerances.

Initial positions are enforced exactly by construction using: $\hat{\vu}(t)=\vu_0 + g(t)\,\tilde{\vu}(t)$ where $g(0)=0$ and $\tilde{\vu}(t)=[\tilde{\rho(t)},\tilde{\theta(t)}]^\top$ is the unconstrained network output. Differentiation shows that the latent output at $t=0$ determines the initial velocity: 
for the angular component, $\tilde{\theta}(0)$ equals $\dot{\theta}(0)$, while for the 
radial component the velocity is proportional to $\tilde{\rho}(0)$, scaled by the 
constant factor $\sigma(\rho_0)$ arising from the softplus derivative. 
Initial velocity is not enforced exactly but is instead penalized during training.

We compare two common gating functions:
an exponential gate $g(t)=1-e^{-t}$ and a linear gate $g(t)=t$. 
Both gating functions satisfy the admissibility conditions of \cite{babni2025error}, 
i.e., $g(0)=0$, $g'(0)\neq 0$, and no additional zeros on $(0,T]$.

To ensure positivity of the radial coordinate, after the gate, we apply
\[
r(t)=r_{\min}+\softplus(\hat{\rho(t)}), \qquad r_{\min}>0.
\]
Here $\softplus(x)=\log(1+e^{x})$ is the smooth softplus function, ensuring
positivity of $r(t)$ while differentiable.

As a baseline, we use a standard fully connected PINN in which
$\tilde{\vu}(t)$ is parameterized by a three-layer MLP with 128 hidden
units per layer and $\tanh$ activations
(33{,}538 trainable parameters).
Spectral models replace the MLP with a Fourier feature map
$\mPhi(t)\in\mathbb{R}^D$ followed by a linear head,
while keeping the same IC embedding and training objective:
\[
\tilde{\vu}(t)=\mW\mPhi(t)+\vb,
\qquad
\mW\in\mathbb{R}^{2\times D},\ \vb\in\mathbb{R}^2.
\]

For both the fixed-frequency and adaptive-frequency Fourier models, we use
32 log-spaced frequencies (two bands: 16 in $[0.5,5.0]$ and 16 in $[5.0,15.0]$), producing $D=64$ 
Fourier features, since each frequency contributes both a $\sin(\omega_i t)$ and a $\cos(\omega_i t)$ term. The fixed Fourier spectral model contains 130 parameters, while the
adaptive Fourier spectral model contains 162 parameters, reflecting the
inclusion of learnable frequency parameters.

\subsection{Gradient scaling induced by IC gating}

Although the choice of $g(t)$ does not change the underlying function class
satisfying the IC constraint, it directly affects optimization through
time-dependent Jacobian scaling. Under IC embedding with $g(t)$ independent of $\theta$,
\[
\hat{\vu}(t) = \vu_0 + g(t)\,\tilde{\vu}(t),
\qquad
\frac{\partial \hat{\vu}(t)}{\partial \theta}
=
g(t)\,\frac{\partial \tilde{\vu}(t)}{\partial \theta}.
\]
Thus, parameter sensitivities are scaled pointwise in physical time by $g(t)$,
inducing an implicit temporal reweighting of gradient propagation.

This effect can be formalized through the Neural Tangent Kernel (NTK)
framework applied to the physics residual loss. Let $R_\theta(t)$ denote the
residual vector and consider the squared residual objective
\[
\mathcal{L}_{\mathrm{phys}}(\theta)
=
\frac{1}{2}
\sum_i
\|R_\theta(t_i)\|^2.
\]
Define $e_i := R_\theta(t_i)$ and linearize around initialization $\theta_0$:
\[
R_\theta(t)
\approx
R_{\theta_0}(t)
+
J_R(t)(\theta-\theta_0),
\qquad
J_R(t)
=
\left.
\frac{\partial R_\theta(t)}{\partial \theta}
\right|_{\theta_0}.
\]
Under continuous-time gradient descent
\(
\frac{d\theta}{d\tau} = -\nabla_\theta \mathcal{L}_{\mathrm{phys}},
\)
we obtain, to first order,
\[
\nabla_\theta \mathcal{L}_{\mathrm{phys}}
\approx
\sum_i J_R(t_i)^\top e_i,
\qquad
\frac{de_i}{d\tau}
\approx
J_R(t_i)\frac{d\theta}{d\tau}
=
-
\sum_j
J_R(t_i)J_R(t_j)^\top e_j.
\]
Thus, the residual errors evolve under an NTK matrix
\[
K_{ij}
=
J_R(t_i)J_R(t_j)^\top.
\]

Because the residual depends on $\hat{\vu}(t)$ and its time derivatives,
the residual Jacobian $J_R(t)$ inherits multiplicative factors of
$g(t)$ (and, for derivative terms, also $g'(t)$ and $g''(t)$). 
A linear gate $g(t)=t$ yields Jacobians that increase with time, emphasizing 
later-time residuals. In contrast, the exponential gate $g(t)=1-e^{-t}$
quickly saturates, producing a more uniform temporal weighting.
Consequently, the effective kernel is reweighted in a time-dependent manner,
altering how gradient descent distributes emphasis across the physical trajectory.

\subsection{Training and Evaluation}

The training objective combines the physics residual loss with a soft
initial-velocity penalty,
\[
\mathcal{L}
=
\lambda_{\mathrm{phys}}\,\mathcal{L}_{\mathrm{phys}}
+
\lambda_{\mathrm{IC}}\,\mathcal{L}_{\mathrm{IC,vel}},
\]
where $\mathcal{L}_{\mathrm{phys}}$ is the mean-squared ODE residual at
collocation points and $\mathcal{L}_{\mathrm{IC,vel}}$ penalizes the induced
initial velocity. We set $\lambda_{\mathrm{phys}}=1$ and vary
$\lambda_{\mathrm{IC}}\in\{0,50\}$.

All models are trained with Adam for 5{,}000 updates using a constant learning
rate of $10^{-3}$ and no weight decay. Collocation points are resampled uniformly
at each iteration with $N_{\mathrm{coll}}=2000$ interior points and
$N_{\mathrm{IC}}=20$ initial-condition points. 
Errors are evaluated on a fixed grid of $N_{\mathrm{eval}}=2000$ points over $t\in[0,T]$. 

Performance is evaluated using the relative $L^2$ error
(ReL2E) and the maximum absolute error (MaxAE). 
Reference solutions are computed using \texttt{solve\_ivp} (DOP853). 
Let the wrapped angular difference and the instantaneous state error be defined as
\[
\Delta\theta(t)=\operatorname{atan2}\!\big(\sin(\theta(t)-\theta_{\mathrm{ref}}(t)),
\cos(\theta(t)-\theta_{\mathrm{ref}}(t))\big), \qquad
\mathbf{e}(t)=
\begin{bmatrix}
r(t)-r_{\mathrm{ref}}(t)\\
\Delta\theta(t)
\end{bmatrix}.
\]
The relative $L^2$ error over the trajectory and maximum absolute error are defined as 
\[ 
\mathrm{ReL2E}_u = \frac{\|\mathbf{e}\|_{L^2}}{\|\mathbf{u}_{\mathrm{ref}}\|_{L^2}}, \qquad \|\mathbf{e}\|_{L^2} = \left( \int_0^T \|\mathbf{e}(t)\|_2^2 \, dt \right)^{1/2}, \mathrm{MaxAE}_u = \max_{t\in[0,T]} \|\mathbf{e}(t)\|_2 .
\] 
where $\|\cdot\|_2$ denotes the Euclidean norm over state components and $\|\cdot\|_{L^2}$ denotes the $L^2$ norm over time. 
Here the denominator normalizes by the $L^2$ norm of the full reference trajectory $\mathbf{u}_{\mathrm{ref}}(t)=[r_{\mathrm{ref}}(t),\theta_{\mathrm{ref}}(t)]^\top$. 

As specified in figure captions, some runs use $20$ seeds, with seeds in $[0,19]$. For 10 random seeds, seeds are in $[0,9]$. For $k=20$ and $k=60$ adaptive $20$ seed runs, paired Wilcoxon signed-rank tests compare gates for each $(k,\text{metric})$ using shared random seeds. Holm--Bonferroni correction is applied to control family-wise error across tests. Error bars show mean $\pm$ 95\% confidence intervals.

\section{Results}

\begin{figure}[H]
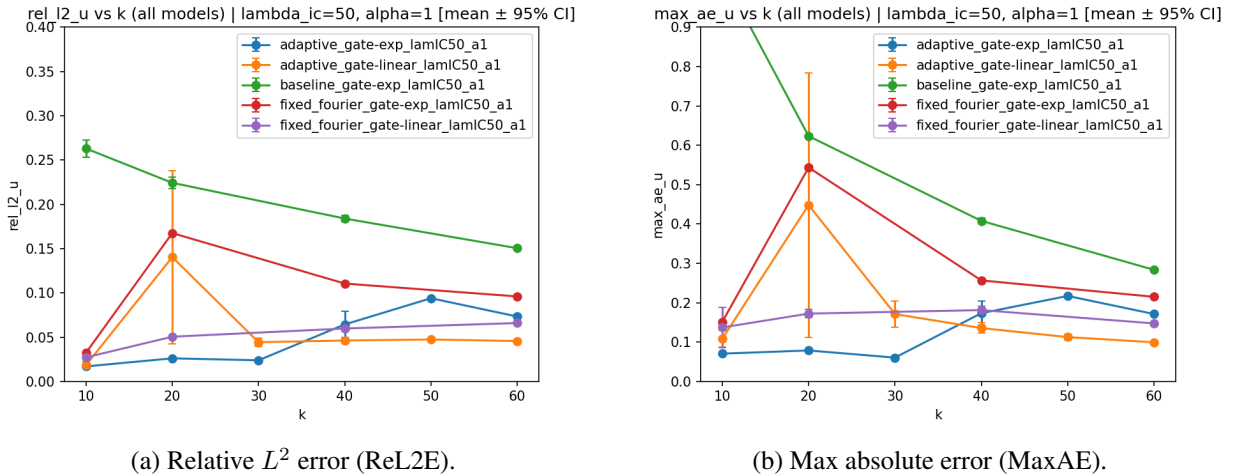

\centering
\begin{subfigure}[t]{0.49\linewidth}
  \centering
  \LamICPlot{lamIC50}{rel_l2_u_vs_k_all_models__zoom_noBaselineLinear}{\linewidth}
  \caption{Relative $L^2$ error (ReL2E).}
\end{subfigure}
\hfill
\begin{subfigure}[t]{0.49\linewidth}
  \centering
  \LamICPlot{lamIC50}{max_ae_u_vs_k_all_models__zoom_noBaselineLinear}{\linewidth}
  \caption{Max absolute error (MaxAE).}
\end{subfigure}
\caption{Gate comparison under $\lambda_{\mathrm{IC}}=50$ (mean $\pm$ 95\% CI).
Adaptive models use 20 seeds for each $k \in \{20, 30, 50, 60\}$ and 10 seeds elsewhere.}
\label{fig:lamIC50_main}
\end{figure}

Figure~\ref{fig:lamIC50_main} summarizes performance across stiffness values under $\lambda_{\mathrm{IC}}=50$. 
Similar trends are observed under $\lambda_{\mathrm{IC}}=0$ (Appendix A.2 Figure ~\ref{fig:lamIC0_main}). 

\FloatBarrier

\FloatBarrier

At moderate stiffness ($k=20$), adaptive spectral models with exponential gating often achieve lower error than
their linear-gated counterparts, although the magnitude of improvement varies
across random seeds and diagnostic plots reveal heterogeneous behavior. At
higher stiffness ($k=60$), linear gating yields consistently lower error for both
ReL2E and MaxAE across all seeds. Intermediate stiffness values (k = 30, 50) show transitional behavior.

Paired Wilcoxon signed-rank tests comparing adaptive exponential and linear gates
at $k=20$ and $k=60$ confirm these trends, with statistically significant
differences after Holm correction (raw and adjusted p-values are reported in Appendix~\ref{app:wilcoxon}, Table~\ref{tab:wilcoxon_gate}). 

The results indicate a stiffness-dependent reversal in the preferred IC gate, consistent with the gradient-scaling analysis introduced in Section~2. Additional, higher stiffness values for $k \in \{70, 80, 90, 100, 110, 120, 130\}$ each with $20$ seeds for the adaptive model are reported in Appendix A.2 and exhibit multiple crossover trends. Baseline models exhibit larger overall gate sensitivity, with a visible performance gap between linear and exponential gating (Appendix A.2 Figure ~\ref{fig:rel_baseline_linear}), whereas adaptive spectral models exhibit a stiffness-dependent crossover in dominance.

\section{Conclusion}

We studied how initial-condition (IC) gating influences optimization in spectral PINNs for stiff nonlinear ODEs. Comparing linear and exponential gates, we observed stiffness-dependent changes in relative dominance: exponential gating performs better at moderate stiffness ($k=20$), while linear gating becomes preferable at higher stiffness ($k=60$), with statistically significant differences under paired Wilcoxon tests. Additional reversals are observed at larger $k$ (Appendix A.2 Figure~\ref{fig:expanded_adaptive}). We study a single stiff nonlinear ODE under fixed training settings; extending the analysis to other systems and gating strategies remains future work.

From a Jacobian perspective, IC gating scales parameter sensitivities by $g(t)$; in the residual NTK formulation, this induces a time-dependent reweighting of the effective kernel through factors of $g(t)$ and its derivatives, thereby modifying gradient-descent conditioning without changing the underlying function class defined by the IC constraint. Baseline models exhibit greater sensitivity to this temporal reweighting (Appendix A.2 Figure~\ref{fig:rel_baseline_linear}), whereas adaptive spectral models display stiffness-dependent changes in relative dominance. IC embeddings should therefore be treated as a first-order design choice in stiff regimes.

\subsection*{Code Availability}

Code for reproducing the experiments is available at: \\
\url{https://github.com/isabelayepes/gradient-scaling-pinns}.

\subsection*{Workshop Appearance}

This work appeared at the ICLR 2026 AI\&PDE Workshop on OpenReview.

\subsection*{Use of Large Language Models}

Large language models were used as an assistive tool during the preparation of
this manuscript, primarily to improve clarity, organization, and LaTeX
formatting, and to support code debugging and experimental scripting. The authors
take full responsibility for the content of the paper and for any errors therein.

\FloatBarrier

\bibliographystyle{plainnat}
\bibliography{references}

\begin{thebibliography}{6}
\providecommand{\natexlab}[1]{#1}
\providecommand{\url}[1]{\texttt{#1}}
\expandafter\ifx\csname urlstyle\endcsname\relax
  \providecommand{\doi}[1]{doi: #1}\else
  \providecommand{\doi}{doi: \begingroup \urlstyle{rm}\Url}\fi

\bibitem[Babni et~al.(2025)Babni, Jamiai, and Rodrigues]{babni2025error}
Atmane Babni, Ismail Jamiai, and José~Alberto Rodrigues.
\newblock Error estimates and generalized trial constructions for solving odes
  using physics-informed neural networks.
\newblock \emph{Mathematical \& Computational Applications}, 30\penalty0
  (6):\penalty0 127, 2025.

\bibitem[Jacot et~al.(2018)Jacot, Gabriel, and Hongler]{jacot2018ntk}
Arthur Jacot, Franck Gabriel, and Cl{\'e}ment Hongler.
\newblock Neural tangent kernel: Convergence and generalization in neural
  networks.
\newblock In \emph{Advances in Neural Information Processing Systems}, 2018.

\bibitem[Lagaris et~al.(1998)Lagaris, Likas, and
  Fotiadis]{lagaris1998artificial}
Isaac~E. Lagaris, Aristidis Likas, and Dimitrios~I. Fotiadis.
\newblock Artificial neural networks for solving ordinary and partial
  differential equations.
\newblock \emph{IEEE Transactions on Neural Networks}, 1998.

\bibitem[Wang et~al.(2020)Wang, Wang, and Perdikaris]{wang2020fourier}
Sifan Wang, Hanwen Wang, and Paris Perdikaris.
\newblock On the eigenvector bias of fourier feature networks.
\newblock \emph{arXiv preprint arXiv:2012.10047}, 2020.

\bibitem[Wang et~al.(2021)Wang, Teng, and Perdikaris]{wang2021whyfail}
Sifan Wang, Yujie Teng, and Paris Perdikaris.
\newblock When and why pinns fail to train: A neural tangent kernel
  perspective.
\newblock \emph{Journal of Computational Physics}, 2021.

\bibitem[Xiong et~al.(2025)Xiong, Zhang, Hu, Gao, and Deng]{xiong2025svsnn}
Xiaodong Xiong, Zhen Zhang, Rui Hu, Cheng Gao, and Zhi Deng.
\newblock Separated-variable spectral neural networks: A physics-informed
  learning approach for high-frequency pdes.
\newblock \emph{arXiv preprint arXiv:2508.00628}, 2025.

\end{thebibliography}

\appendix
\section{Appendix}

\FloatBarrier

\subsection{Tables}
\label{app:wilcoxon}

\begin{table}[t]
\centering
\caption{Paired Wilcoxon signed-rank tests across seeds comparing IC gates for the adaptive model. Lower is better for both metrics. Winner indicates whether $A$ or $B$ attains lower error. We report both raw two-sided p-values and Holm-adjusted two-sided p-values; Holm correction is applied across the tested $(k,\text{metric})$ pairs within each setting.}
\label{tab:wilcoxon_gate}

\small
\setlength{\tabcolsep}{4pt}
\begin{tabular}{l c l c c c c c c c c}
\toprule
Setting & $k$ & Metric & Winner & $n$ & mean($A$) & mean($B$) & frac(win) & $p_{\mathrm{raw}}$ & $p_{\mathrm{Holm}}$ & $p_{\mathrm{1s}}$ \\
\midrule
\multicolumn{11}{l}{\small noIC: $A$=adaptive\_gate-exp\_lamIC0\_a1, $B$=adaptive\_gate-linear\_lamIC0\_a1} \\
noIC & 20 & max\_ae\_u & $A \downarrow$ & 20 & 0.139 & 0.519 & 0.85 & 0.0012 & 0.0020 & 6.0e-04 \\
noIC & 20 & rel\_l2\_u & $A \downarrow$ & 20 & 0.049 & 0.163 & 0.85 & 0.0010 & 0.0020 & 5.1e-04 \\
noIC & 60 & max\_ae\_u & $B \downarrow$ & 20 & 0.180 & 0.088 & 1.00 & $<10^{-4}$ & $<10^{-4}$ & $<10^{-4}$ \\
noIC & 60 & rel\_l2\_u & $B \downarrow$ & 20 & 0.082 & 0.045 & 1.00 & $<10^{-4}$ & $<10^{-4}$ & $<10^{-4}$ \\
\multicolumn{11}{l}{\small lIC50: $A$=adaptive\_gate-exp\_lamIC50\_a1, $B$=adaptive\_gate-linear\_lamIC50\_a1} \\
lIC50 & 20 & max\_ae\_u & $A \downarrow$ & 20 & 0.079 & 0.448 & 0.85 & $<10^{-4}$ & $<10^{-4}$ & $<10^{-4}$ \\
lIC50 & 20 & rel\_l2\_u & $A \downarrow$ & 20 & 0.026 & 0.140 & 1.00 & $<10^{-4}$ & $<10^{-4}$ & $<10^{-4}$ \\
lIC50 & 60 & max\_ae\_u & $B \downarrow$ & 20 & 0.171 & 0.099 & 1.00 & $<10^{-4}$ & $<10^{-4}$ & $<10^{-4}$ \\
lIC50 & 60 & rel\_l2\_u & $B \downarrow$ & 20 & 0.073 & 0.046 & 1.00 & $<10^{-4}$ & $<10^{-4}$ & $<10^{-4}$ \\\bottomrule
\end{tabular}
\end{table}

\subsection{Additional Figures}

\begin{figure}[t]
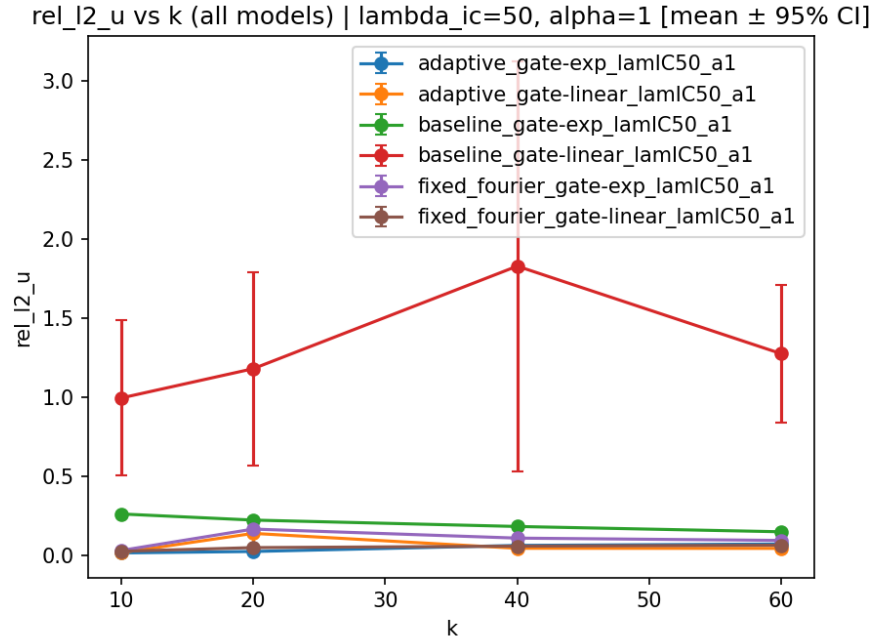

\centering
\LamICPlot{lamIC50}{rel_l2_u_vs_k_all_models}{0.75\linewidth}
\caption{Full-scale ReL2E versus stiffness for $\lambda_{\mathrm{IC}}=50$.}
\label{fig:rel_baseline_linear}
\end{figure}

\begin{figure}[t]
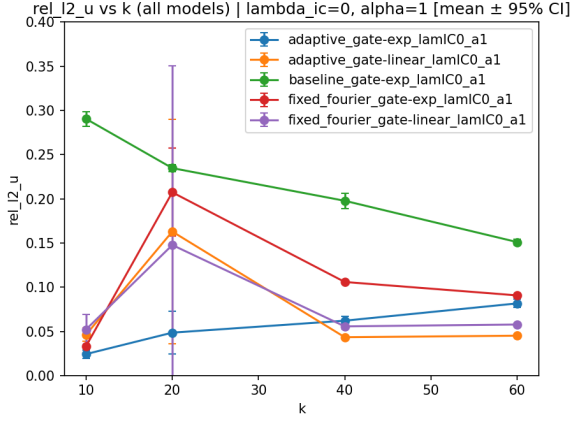
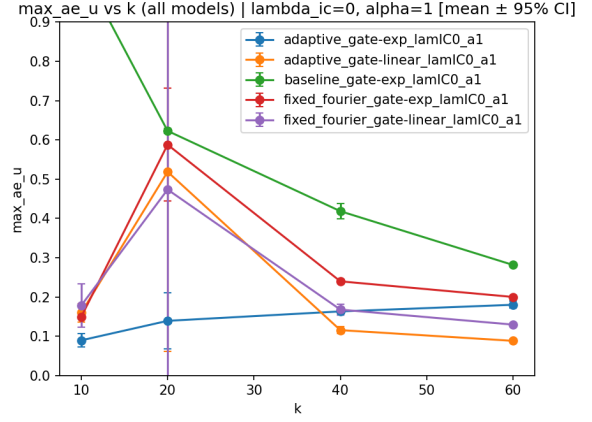

\centering
\begin{subfigure}[t]{0.49\linewidth}
  \centering
  \LamICPlot{lamIC0}{rel_l2_u_vs_k_all_models__zoom_noBaselineLinear}{\linewidth}
  \caption{Relative $L^2$ error (ReL2E).}
\end{subfigure}
\hfill
\begin{subfigure}[t]{0.49\linewidth}
  \centering
  \LamICPlot{lamIC0}{max_ae_u_vs_k_all_models__zoom_noBaselineLinear}{\linewidth}
  \caption{Max absolute error (MaxAE).}
\end{subfigure}
\caption{Gate comparison under $\lambda_{\mathrm{IC}}=0$ (mean $\pm$ 95\% CI).
Adaptive models use 20 seeds at $k = 20$ and $k = 60$ (used for statistical testing) and 10 seeds elsewhere.}
\label{fig:lamIC0_main}
\end{figure}

\begin{figure}[t]
\centering
\begin{subfigure}[t]{0.49\linewidth}
  \centering
  \LamICPlot{expanded_adaptive}{rel_l2_u_vs_k_all_models__zoom_spectralOnly}{\linewidth}
  \caption{Relative $L^2$ error (ReL2E).}
\end{subfigure}
\hfill
\begin{subfigure}[t]{0.49\linewidth}
  \centering
  \LamICPlot{expanded_adaptive}{max_ae_u_vs_k_all_models__zoom_spectralOnly}{\linewidth}
  \caption{Max absolute error (MaxAE).}
\end{subfigure}
\caption{Gate comparison under $\lambda_{\mathrm{IC}}=50$ (mean $\pm$ 95\% CI).
Adaptive models use 20 seeds for each $k \in \{20, 30, 50, 60, 70, 80, 90, 100, 110, 120, 130\}$ and 10 seeds elsewhere.}
\label{fig:expanded_adaptive}
\end{figure}
\FloatBarrier

\end{document}